\documentclass[letterpaper, 10 pt, conference]{ieeeconf}  

\IEEEoverridecommandlockouts                              

\usepackage{graphics} 
\usepackage{graphicx}
\usepackage{subcaption}
\usepackage{tikz}
\usepackage{amsmath} 
\usepackage{amssymb}  
\usepackage{multirow}
\usepackage{cite}
\usepackage{threeparttable}
\usepackage{xcolor} 
\usepackage{cleveref} 
\usepackage{url} 
\usepackage{hhline}
\usepackage{tabularx}

\newcommand{\mbf}[1]{\mathbf{#1}}

\newcommand{\mbfhat}[1]{\hat{\mbf{#1}}}

\newcommand{\mbs}[1]{\boldsymbol{#1}}

\DeclareMathAlphabet{\mbfh}{OML}{cmm}{b}{it}

\newcommand{\cframe}[1]{\ensuremath \underrightarrow{\mathcal{F}}_{#1}}

\newcommand{\argmin}{\operatorname*{argmin}}

\newcommand{\pd}[2]{\dfrac{\partial #1}{\partial #2}}

\newcommand{\bbm}{\begin{bmatrix}}
\newcommand{\ebm}{\end{bmatrix}}

\newcommand\T{\rule{0pt}{2.6ex}}        
\newcommand\B{\rule[-1.2ex]{0pt}{0pt}}  
\title{\LARGE\bf Entropy-Based \textit{Sim}(3) Calibration of 2D Lidars to Egomotion Sensors}
\author{Jacob Lambert$^1$, Lee Clement$^1$, Matthew Giamou$^2$, and Jonathan Kelly$^1$
\thanks{$^{1}$J. Lambert, L. Clement, and J. Kelly are with the Space and Terrestrial Autonomous Robotic Systems Laboratory, Institute for Aerospace Studies, University of Toronto. {\tt\small \{jacob.lambert, lee.clement\}@mail.utoronto.ca, j.kelly@utias.utoronto.ca}}
\thanks{$^{2}$M. Giamou is with the Laboratory for Information and Decision Systems, Massachusetts Institute of Technology. {\tt \small mgiamou@mit.edu}}
}

\begin{document}

\maketitle

\begin{abstract}
This paper explores the use of an entropy-based technique for point cloud reconstruction with the goal of calibrating a lidar to a sensor capable of providing egomotion information. We extend recent work in this area to the problem of recovering the \textit{Sim}(3) transformation between a 2D lidar and a rigidly attached monocular camera, where the scale of the camera trajectory is not known \emph{a priori}. We demonstrate the robustness of our approach on realistic simulations in multiple environments, as well as on data collected from a hand-held sensor rig. Given a non-degenerate trajectory and a sufficient number of lidar measurements, our calibration procedure achieves millimetre-scale and sub-degree accuracy. Moreover, our method relaxes the need for specific scene geometry, fiducial markers, or overlapping sensor fields of view, which had previously limited similar techniques.
\end{abstract}

\section{INTRODUCTION}

Multisensor payloads have become the norm in many mobile robotic systems. 
In particular, lidars are frequently used on multisensor platforms that must estimate distances accurately. While visual sensors such as stereo cameras can recover this information using multiple view geometry, they do not have the accuracy of typical lidars for depth estimation \cite{strecha2007multi}. Inexpensive RGB-D cameras are also commonly used for building dense, coloured 3D maps, but they have limited range and are generally too sensitive to lighting conditions to be used outdoors. In fact, modern 3D lidars are commonly employed to generate ground truth depth maps against which to evaluate the performance of visual estimation techniques \cite{Geiger2013IJRR}.  The range and lighting-invariance of a lidar makes it suitable for both indoor and outdoor operation; lidars have proven to be an unparalleled sensor choice for unmanned ground vehicles such as self-driving cars\cite{WuROBIO2015, SugerICRA2015}. 

\begin{figure}[ht!]
    \centering
    \begin{subfigure}[b]{0.95\linewidth}
        \begin{tikzpicture}[      
        every node/.style={anchor=south west,inner sep=0pt},
        x=1mm, y=1mm,
      ]   
     \node (fig1) at (0,0)
       {\includegraphics[width=0.95\linewidth]{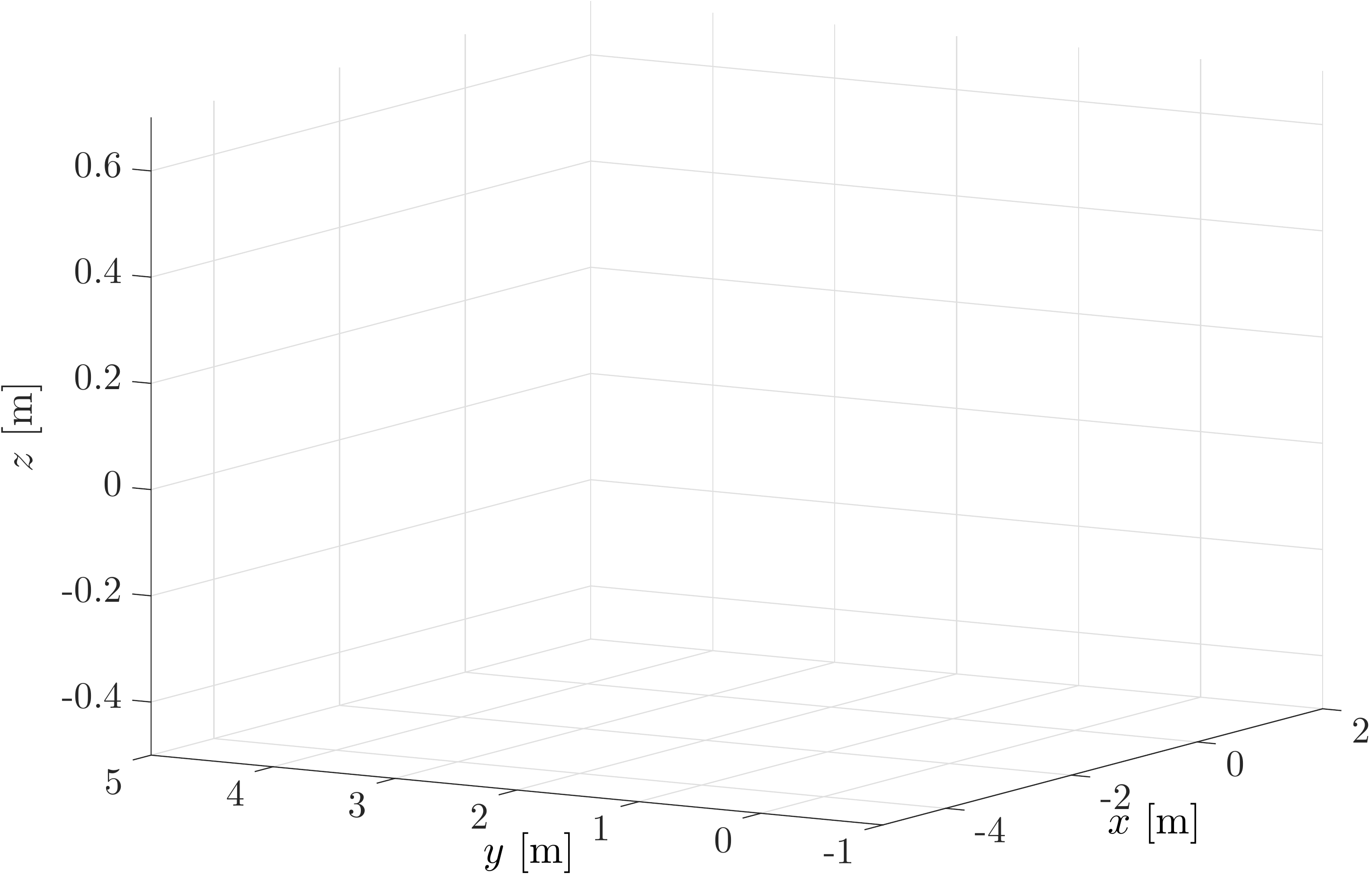}};
    \node (fig2) at (8,6)
       {\includegraphics[keepaspectratio=false,width=0.75\linewidth, height=0.5\linewidth]{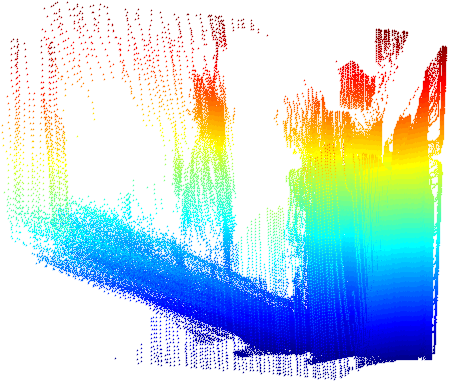}};  
		\end{tikzpicture}
        \caption{Point cloud before calibration.}
        \label{fig:pre_calib}
    \end{subfigure}
     
     \begin{subfigure}[b]{0.95\linewidth}
        \begin{tikzpicture}[      
        every node/.style={anchor=south west,inner sep=0pt},
        x=1mm, y=1mm,
      ]   
     \node (fig1) at (0,0)
       {\includegraphics[width=0.95\linewidth]{figs/point_cloud_background}};
     \node (fig2) at (8.5,7.5)
       {\includegraphics[keepaspectratio=false,width=0.8\linewidth, height=0.55\linewidth]{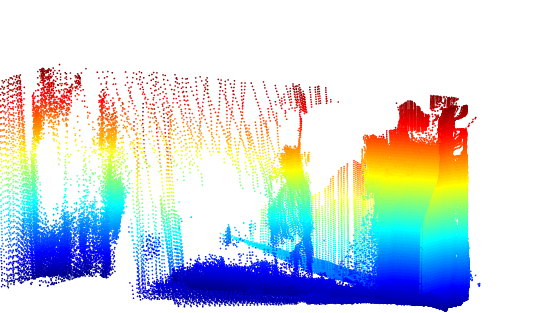}};  
		\end{tikzpicture}
        \caption{Point cloud after calibration using our method.}
        \label{fig:post_calib}
    \end{subfigure}
    \caption{\small Visualization of a point cloud collected in the environment pictured in \Cref{fig:lids_environment}, using initial values and mean values from \Cref{tab:real_results}. Note the smooth wall segments on the right side of \Cref{fig:post_calib}.
    }
    \label{fig:point_clouds}
    \vspace{-10pt}
\end{figure}

Robotic platforms will often include both cameras and lidars as complementary sensors. Cameras provide dense visual information, but in the form of 2D images which complicates the recovery of 3D information. Moreover, in the case of monocular motion estimation, a scale ambiguity exists: metric scale cannot be recovered from monocular images alone. In contrast, lidars provide precise range information, ideal for creating depth maps and naturally complementing camera data in scene reconstruction \cite{miksikIROS2015}. 

In general, successfully combining information from multiple sensors requires knowledge of the coordinate transformations between them---an inaccurate transform leads to systematic error in the fused sensor data. Manually measuring the translational and rotational offset between sensors is inaccurate and may be impractical since the origins of the sensor coordinate systems are often embedded within the sensors themselves. Instead, we would like to use data-driven techniques to estimate the sensor-to-sensor transformations. Conventional approaches to this problem often require a specific calibration procedure with a target (e.g., a checkerboard in the case of cameras), although techniques have been developed recently for automatic calibration ``in the wild''. Despite these advances, current methods show a lack of generality in that they are often developed for specific sensor suites, require specialized scene structure, or place restrictions on the sensors' fields of view (FOVs). These limitations are especially problematic for long-term robotic deployments, during which the extrinsic calibration parameters may slowly drift or suddenly change; a calibration procedure applicable to an unknown environment and sensor configuration would then be necessary.

In this work, we seek to address these limitations by developing a generalized technique for the extrinsic calibration of a 2D lidar to a sensor capable of providing egomotion information. Our method builds directly upon previous work \cite{Jian2005ICCV,MaddernICRA2012} by representing the lidar point cloud as a Gaussian Mixture Model (GMM) and minimizing its R\'enyi Quadratic Entropy (RQE) in order to recover the coordinate transformation between the two sensors. In contrast to \cite{Jian2005ICCV,MaddernICRA2012}, which are concerned with estimating \textit{SE}(3) rigid body transformations between sensor pairs, we estimate \textit{Sim}(3) \emph{similarity transformations}. This allows us to use a monocular camera as the egomotion sensor, despite the scale ambiguity in its trajectory reconstruction. The scale factor is estimated as part of the calibration process. 

The remainder of the paper is organized as follows: \Cref{sec:related} situates our paper with respect to existing literature. \Cref{sec:theory,sec:algorithm} lay out the mathematical underpinnings of our approach as well as practical considerations in its implementation. In \Cref{sec:sim}, we show through simulations that we can recover the \textit{Sim}(3) transformation between a lidar and a noisy egomotion sensor. Experiments presented in \Cref{sec:exp} demonstrate that our technique can also recover the absolute scale of a trajectory estimated from a monocular image stream. We show that our technique performs reliably even in environments with complex structures. Finally, we conclude by motivating the development of a spatiotemporal calibration technique based on entropy minimization.

\section{RELATED WORK}\label{sec:related}

Extrinsic sensor calibration is an active area of research and a variety of approaches have been developed. The classic approach for extrinsic calibration involves a laboratory procedure making use of a known calibration object. For cameras and 2D lidars, a common choice is to identify lines on a planar calibration target \cite{2011_Kwak_Extrinsic,2011_Naroditsky_Automatic,2013_Guo_Analytical}. More complex patterned planes have also been used in some work \cite{2009_Willis_Linear,2013_Carrasco-Ochoa_LIDAR}. 

As a step towards removing the need for specific calibration objects, several methods instead attempt to identify appropriate calibration geometry (`features') in the environment. For example, surface normals can be extracted from lidar data and processed for comparison with camera images \cite{2015_Pandey_Automatic,2012_Taylor_Mutual}. Picking out linear features, typically lines formed at the intersection of planar surfaces, is a practical option in indoor environments \cite{2012_Yang_Simple,2013_Moghadam_Line-based}. Note, however, that all the methods mentioned thus far require the camera and lidar to simultaneously observe the same features, which requires an overlapping FOV between the two sensors. 

Some recent techniques relax the overlapping FOV requirement, by allowing each sensor to observe the same features at different times \cite{2014_Angelats_One,2016_Bok_Extrinsic,2013_Napier_Cross-Calibration}. These approaches also require accurate temporal alignment of the sensor data. While spatiotemporal calibration techniques are few, the case of non-overlapping lidar and stereo cameras has been addressed by identifying and matching planar surfaces\cite{2014_Rehder_Spatio-Temporal}. The method of \cite{2014_Rehder_Spatio-Temporal} represents a significant step towards generalized spatiotemporal calibration, but requires specific features and metrically accurate trajectory estimates.

In contrast to previous work, our technique (an extension of \cite{Sheehan2010} and \cite{MaddernICRA2012}) requires neither overlapping FOVs, nor observations of specific features in the environment. We assume only a non-degenerate sensor trajectory\cite{brookshire2013extrinsic} and a scene with a certain level of structure, as discussed in \Cref{sec:conc}. Metric accuracy is not necessary, as we explicitly optimize for scale.

\section{THEORY}\label{sec:theory}

\subsection{Problem Formulation}
For our particular application, we assume that the camera's intrinsic parameters have been calibrated and that a monocular motion estimation system (e.g. ORB-SLAM~\cite{MurArtal:2015:TRO}) provides a reasonable sensor trajectory, accurate up to a scale factor. We stress that this method is not restricted to monocular cameras; any sensor capable of providing egomotion estimates (i.e., the base sensor) could be used in place of the camera.

The base sensor provides a set of $K$ 6DOF poses, $\mbf{Y}$:
\begin{equation}
 \mbf{Y}=\{ \mbf{y}_1, \mbf{y}_2, ..., \mbf{y}_K \},~~ \mbf{y}_k = \left[ x_k~y_k~z_k~\phi_k~\theta_k~\psi_k \right]^T,
\end{equation}
where $\phi$, $\theta$ and $\psi$ are roll, pitch and yaw Euler angles, respectively. Each pose $\mbf{y}_k$ has an associated timestamp $t_k$ and pose covariance matrix $\mbf{Q}_k$. The lidar provides a set of $K\times N$, 2D observations, $\mbf{Z}$, where: 
\begin{equation} 
 \mbf{Z} = \{ \mbf{z}_1, \mbf{z}_2,...,\mbf{z}_K \},~~ \mbf{z}_k = \{ \mbf{z}^{(1)}_k,\mbf{z}^{(2)}_k, ..., \mbf{z}^{(N)}_k\},
\end{equation}
\begin{equation}
 \mbf{z}^{(n)}_k = \left[x^{(n)}_k~y^{(n)}_k\right]^T,
\end{equation}
and each point $\mbf{z}^{(n)}_k$ has associated timestamp $t_k$. We assume that the camera and lidar measurements are temporally aligned and explore the consequences of this assumption in \Cref{sec:sim_td}. For convenience, we adopt a homogeneous representation and express each point in lidar frame $\cframe{L_k}$ as
 \vspace{-3pt}
\begin{equation}
 \mbf{p}^{(n)}_{L_k} = \left[x^{(n)}_k~y^{(n)}_k~0~1\right]^T.
\end{equation}

\noindent Likewise, we express each camera pose as a homogeneous transformation from camera frame $\cframe{C_k}$ to a fixed global frame $\cframe{G}$, where matrix $\mbf{T}_{G,C_k}$ is the homogeneous representation of pose $\mbf{y}_k$.

Our goal is to estimate the set of (constant) transform parameters from the lidar frame $\cframe{L}$ to the camera frame $\cframe{C}$
\begin{equation}
  \mbs{\Xi} = \left[x_L~~y_L~~z_L~~\phi_L~~\theta_L~~\psi_L~~s \right]^T,
\end{equation}
which we use to form the rigid body transformation matrix $\mbf{T}_{{C},{L}}$ from $\cframe{L}$ to $\cframe{C}$. 
The scale factor, $s$, is applied in the transformation from the camera frame to global frame. This defines our inverse sensor model, with which we can estimate the position of a lidar point in the global frame:
\begin{equation}
 \hat{\mbf{p}}^{(n)}_{G,k} = h^{-1}(\mbf{p}^{(n)}_{L_k} \mid \mbf{y}_k,  \mbs{\Xi}) = \mbf{T}_{G,C_k} \mbf{T}_{{C_k},{L_k}} \mbf{p}^{(n)}_{L_k}.
\end{equation}
Omitting the homogeneous component so that $\hat{\mbf{x}}^{(n)}_{G,k}\leftarrow \hat{\mbf{p}}^{(n)}_{G,k}$, we use the Jacobian of this model and the camera pose covariances to obtain a covariance estimate of points in the world frame:
\begin{equation}\label{eq:jacobian}
 \mbs{\Sigma}^{(n)}_k = \mbf{J}^{(n)}_k \mbf{Q}_k {\mbf{J}^{(n)}_k}^T,~~\mbf{J}^{(n)}_k = \pd{h^{-1}(\mbf{x}^{(n)}_{L_k} | \mbf{y}_k,  \mbs{\Xi})}{\mbf{y}_k}.
\end{equation}
Finally, we obtain a set of 3D points $\hat{\mbf{x}}^{(n)}_{G,k} \in \hat{\mbf{X}}$ expressed in the global frame, each with an associated 3 $\times$ 3 covariance matrix $\mbs{\Sigma}^{(n)}_{k}$ and timestamp $t^{(n)}_k$.

\subsection{Point Cloud R\'enyi Quadratic Entropy}
As in \cite{MaddernICRA2012}, we minimize the R\'enyi Quadratic Entropy (RQE)~\cite{renyi1961} in order to maximize the compactness or `crispness' of the estimated point cloud $\hat{\mbf{X}}$. The intuition behind maximizing point cloud compactness follows from the assumption that the surfaces in the environment are structured 2D manifolds as opposed to diffuse, random elements. This assumption holds particularly well in many urban and natural scenes.

The RQE of the continuous random variable $X$ with probability distribution $p(\mathbf{x})$ is defined as:
\begin{equation}\label{eq:RQE}
H[X] = -\log \int p(\mbf{x})^2 d \mbf{x}.
\end{equation}
We represent our point cloud $\hat{\mbf{X}}$ as a Gaussian mixture model (GMM), where $p(\mbf{x})$ is the probability of a lidar measurement being sampled at position $\mbf{x}$ given centroids $\{\hat{\mbf{x}}_1, ..., \hat{\mbf{x}}_M\} \in \hat{\mbf{X}}$:
\begin{equation}\label{eq:GMM}
p(\mbf{x}) = \frac{1}{M} \sum_{i=1}^{M} \mathcal{N}  (\mbf{x}-\hat{\mbf{x}}_i, \mbs{\Sigma}_i + \sigma^2 \mbf{I}),
\end{equation}
where the covariance includes pose uncertainty $\mbf{\Sigma}_i$ calculated using \Cref{eq:jacobian}, and measurement uncertainty in the isotropic kernel $\sigma^2 \mbf{I}$. Note that $M$ may be less than $K \times N$ due to missing lidar measurements (e.g., out of range readings). An analytic representation of the RQE of the GMM can be derived as
\begin{align}
\hspace{-0.4em} H[\hat{\mbf{X}}] =& -\log \int \biggl( \frac{1}{M} \sum_{i=1}^M \mathcal{N}(\mbf{x}-\hat{\mbf{x}}_i, \mbs{\Sigma}_i + \sigma^2 \mbf{I}) \biggr)^2 d\mbf{x} \\
    =& -\log \biggl( \frac{1}{M^2} \sum_{i=1}^M \sum_{j=1}^M \int \mathcal{N}(\mbf{x}-\hat{\mbf{x}}_i, \mbs{\Sigma}_i + \sigma^2 \mbf{I}) \\
    &~~~~~~~~~~~~~~~~~~~~~~~~\mathcal{N}(\mbf{x}-\hat{\mbf{x}}_j, \mbs{\Sigma}_j + \sigma^2 \mbf{I}) d\mbf{x} \biggr) \nonumber \\
    =& -\log \biggl( \frac{1}{M^2} \sum_{i=1}^M \sum_{j=1}^M \mathcal{N}(\hat{\mbf{x}}_i-\hat{\mbf{x}}_j, \mbs{\Sigma}_i + \mbs{\Sigma}_j + 2\sigma^2 \mbf{I})\biggr). \label{eq:double}
\end{align}
Therefore, to calculate the entropy of a GMM, we must compute all pairwise entropy contributions. 

\section{Algorithm} \label{sec:algorithm}

Using \Cref{eq:double}, we wish to solve for the optimal set of calibration parameters $\mbs{\Xi}^*$ which minimizes the total point cloud entropy. For optimization, the cost function can be simplified greatly. First, we can trivially remove the monotonic logarithm function and constant $1/M^2$ factor. Next, we note that for each pair of points $\left( \mbfhat{x}_i, \mbfhat{x}_j \right) \in \mbfhat{X}$, we are calculating their pairwise entropy contribution twice, as seen in the indices of the sums in \Cref{eq:double}. We can therefore solve the equivalent problem\cite{MaddernICRA2012}:
\begin{equation}
\mbs{\Xi}^* = \argmin_{\mbs{\Xi}}~ - \sum_{i=1}^M \sum_{j=i}^M \mathcal{N}(\hat{\mbf{x}}_i-\hat{\mbf{x}}_j, \mbs{\Sigma}_i + \mbs{\Sigma}_j + 2\sigma^2 \mbf{I}),
\end{equation}\label{eq:cost_function}

\noindent noting that the coordinates of each point $\hat{\mbf{x}}_{i}, \hat{\mbf{x}}_{j}$ are functions of the calibrations parameters and the camera poses.


Moreover, a large majority of point pairs contribute negligible entropy. We can avoid unnecessary computation by setting a conservative upper bound on the pairwise distance:
\begin{equation*}
\mathcal{N}(\hat{\mbf{x}}_i-\hat{\mbf{x}}_j, \mbs{\Sigma}_i + \mbs{\Sigma}_j + 2\sigma^2 \mbf{I}) \approx 0~~\mathrm{if}
\end{equation*}
\begin{equation}\label{eq:kfactor}
||\hat{\mbf{x}}_i-\hat{\mbf{x}}_j || \geq 2k \left(\mathrm{max}\left(\lambda_1(\mbs{\Sigma}_i),\lambda_1(\mbs{\Sigma}_j)\right)+\sigma^2\right)
\end{equation}
where $\lambda_1(\mbs{\Sigma})$ is the largest eigenvalue of matrix $\mbs{\Sigma}$. The tuning parameter $k$ allows a trade-off between computation time and cost function accuracy, in comparison to the full O($M^2$) computation.  

Presently, we search for the optimal calibration parameters using a sequence of gradient-free optimizers, first performing a coarse global optimization using the control random search algorithm\cite{Kaelo2006}, followed by a fine local optimization using the Nelder-Mead algorithm\cite{Nelder1965}.

\section{SIMULATIONS}\label{sec:sim}
\begin{table*}
    \centering
    \caption{Average absolute error over ten unique trajectories, for the five simulation environments.}
        \begin{tabular}{r *{8}{c}}
            && \multicolumn{7}{c}{Average absolute error -- $\mu$ ($\sigma$)} \T\B \\ \cline{3-9}
            Environment && $x$ $\left[\mathrm{mm}\right]$ & $y$ $\left[\mathrm{mm}\right]$ & $z$ $\left[\mathrm{mm}\right]$ & $\phi$ [deg] & $\theta$ [deg] & $\psi$ [deg] & Scale $[\times 10^{-3}]$ \T\B \\ \hline
	    Simple Room             && 2.8 (2.5) & 3.1 (2.5) & 5.2 (3.5) & 0.22 (0.12) & 0.051 (0.043) & 0.24 (0.15) & 0.33 (0.30) \T \\
	    Underground Parking Lot && 4.5 (4.4) & 4.8 (4.1) & 5.2 (4.5) & 0.37 (0.22) & 0.11 (0.11) & 0.37 (0.17) & 1.2 (0.9) \\
	    Plane City              && 4.1 (2.1) & 5.2 (4.2) & 4.0 (3.9) & 0.38 (0.23) & 0.18 (0.06) & 0.35 (0.23) & 0.69 (0.55) \\
	    Quadratic Forest        && 4.6 (2.6) & 3.9 (1.6) & 2.9 (3.3) & 0.32 (0.20) & 0.074 (0.54) & 0.35 (0.27) & 0.73 (0.32) \\
	    Triangle Array          && 3.0 (1.9) & 2.9 (1.6) & 4.6 (3.5) & 0.64 (0.60) & 0.10 (0.07) & 0.61 (0.58) & 0.47 (0.26)\B \\ \hline
        \end{tabular}
    \label{tab:sim_results}
\end{table*}
\begin{figure*}
	\centering
	\begin{subfigure}[b]{0.3\textwidth}
    	\centering
		\includegraphics[width=0.9\textwidth]{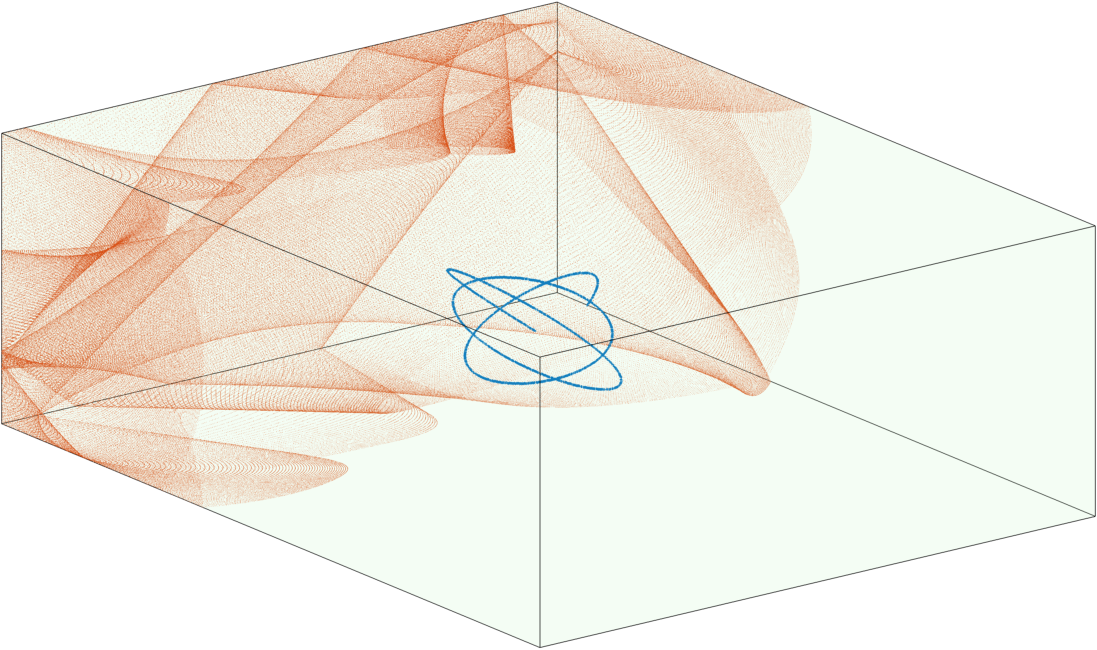}
        \caption{Simple Room}
        \label{fig:simpleroom}
	\end{subfigure}
    ~
    \begin{subfigure}[b]{0.3\textwidth}
    	\centering
		\includegraphics[width=0.9\textwidth]{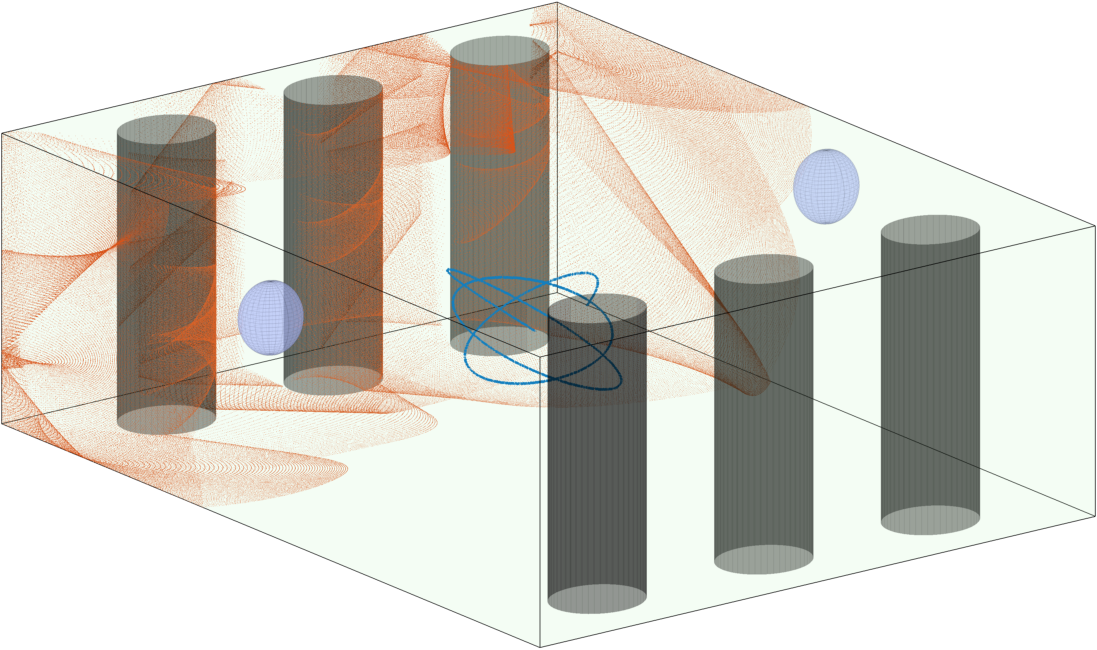}
        \caption{Underground Parking Lot}
        \label{fig:parking}
	\end{subfigure}
    ~
    \begin{subfigure}[b]{0.3\textwidth}
    	\centering
		\includegraphics[width=0.9\textwidth]{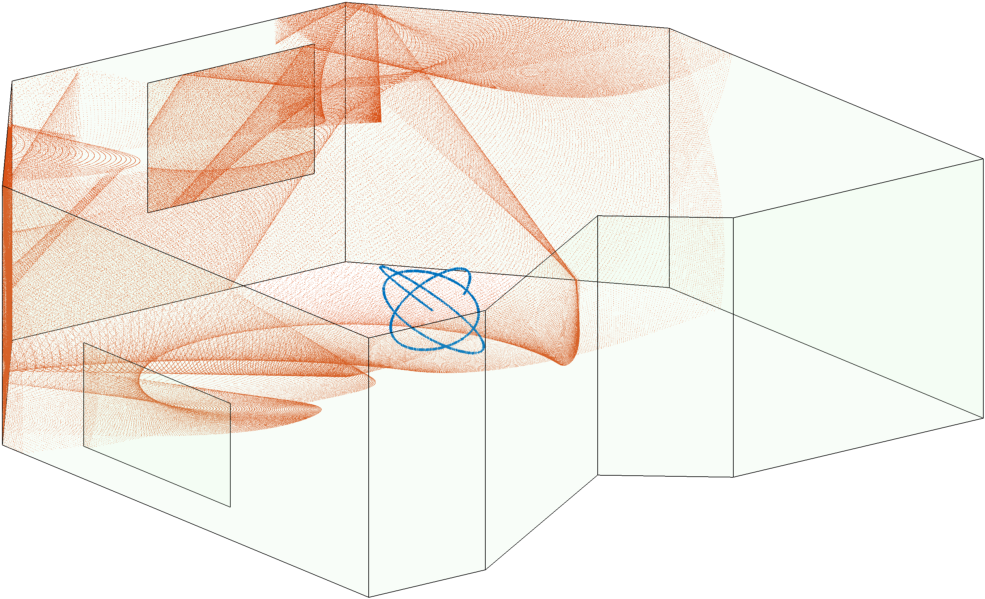}
        \caption{Plane City}
        \label{fig:planecity}
	\end{subfigure}
    
    \begin{subfigure}[b]{0.32\textwidth}
    	\centering
		\includegraphics[width=\textwidth]{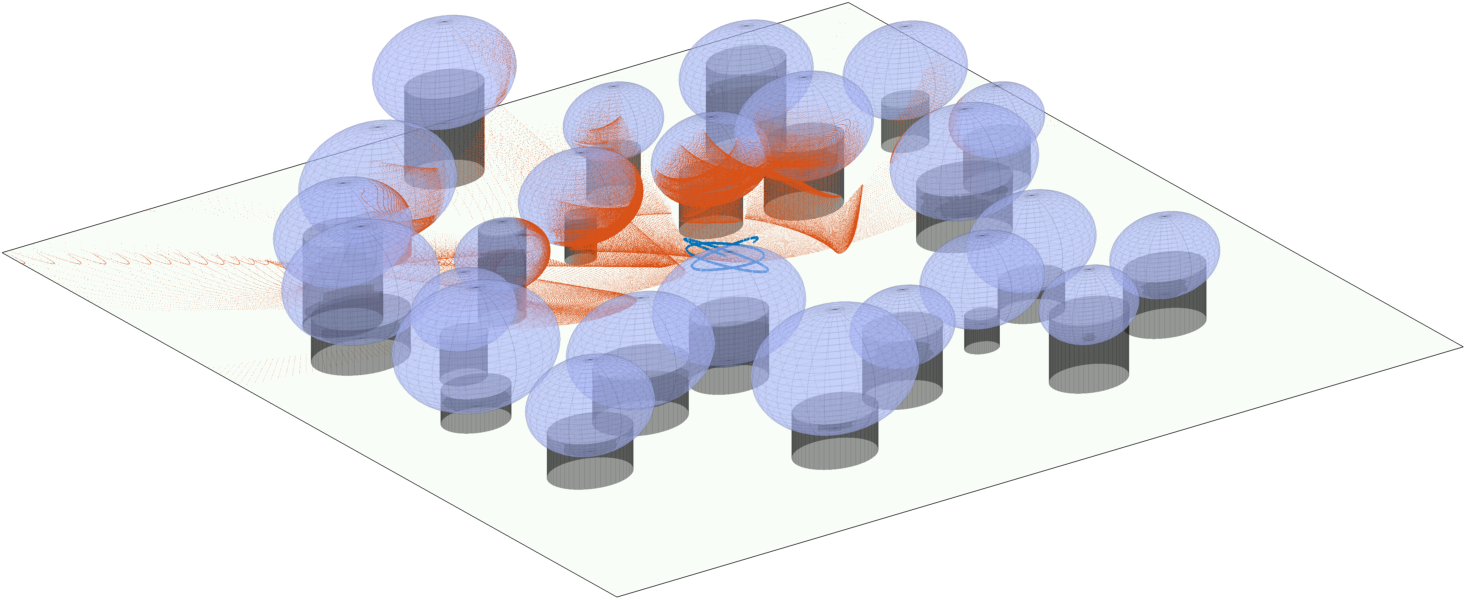}
        \caption{Quadratic Forest}
        \label{fig:forest}
	\end{subfigure}
    ~
    \begin{subfigure}[b]{0.32\textwidth}
    	\centering
		\includegraphics[width=\textwidth]{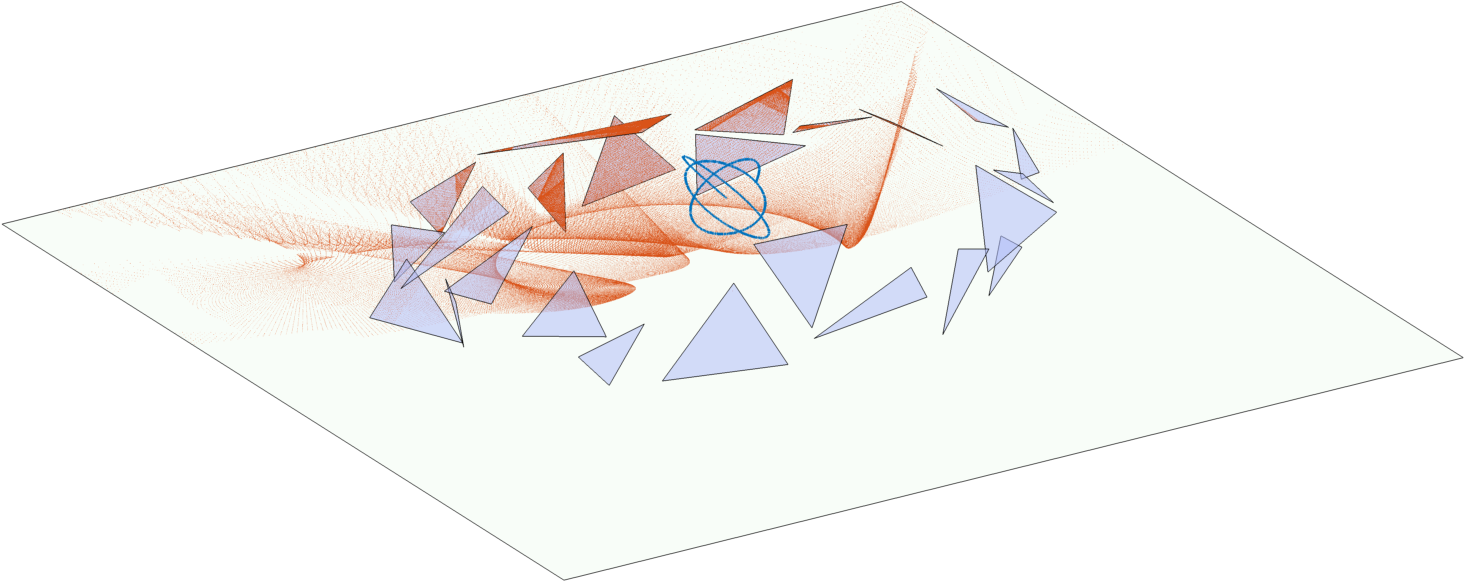}
        \caption{Triangle Array}
        \label{fig:trigeddon}
	\end{subfigure}
    \caption{Simulation environments used to validate our method, with a sample sensor trajectory (blue) and a cut away view of a portion of the lidar point cloud (red).}
\label{fig:sim_envs}
\vspace{-2mm}
\end{figure*}
\subsection{Experimental Setup}
In order to validate our approach, we simulated a 2D lidar rigidly attached to a base sensor, with the base sensor following a known trajectory. To mimic realistic egomotion measurements, uncorrelated zero-mean Gaussian noise with standard deviations of $50~\text{mm}$ and $1^{\circ}$ (translation and rotation, respectively) was added to each pose of the base sensor. Zero-mean Gaussian noise with a $ 50~\text{mm}$ standard deviation was also added to each lidar range measurement. The simulated lidar operated at 40 Hz, with a field of view of $240^{\circ}$ and angular resolution of $0.25^{\circ}$ degrees per beam. The simulated datasets were 50 seconds in duration and contained up to 1.9 million lidar points.

The base sensor trajectories generated for the environments shown in \Cref{fig:sim_envs} involve sinusoidal motions with unique frequencies and amplitudes for each translational and rotational parameter. A geometrically plausible initial trajectory was determined for each environment, and new trajectories were then produced by randomly varying the frequency and amplitude of each sinusoidal component.

\subsection{Simulation Environments}\label{sec:sim_env}
We created five distinct simulation environments, designed to be of increasing difficulty for the algorithm (\Cref{fig:sim_envs}):
\begin{enumerate}
    \item \textbf{Simple Room} consists only of orthogonal planar surfaces where the sensors move inside an enclosed rectangular `room' (\Cref{fig:simpleroom});
    \item \textbf{Underground Parking Lot} adds several pillars (cylinders) to the \emph{Simple Room} environment (\Cref{fig:parking});
    \item \textbf{Plane City} contains planes of varying size, some occluding certain parts of the scene (\Cref{fig:planecity});
    \item \textbf{Quadratic Forest} is an open environment with spheres mounted on top of cylinders, and where the only planar surface is the ground (\Cref{fig:forest}); and
    \item \textbf{Triangle Array} is an open environment filled with non-intersecting triangles of various sizes (\Cref{fig:trigeddon}).
\end{enumerate}

\subsection{Cost Function Validation}
\begin{figure*}
\centering
\begin{subfigure}[t]{0.3\textwidth}
    \centering
    \includegraphics[width=\textwidth]{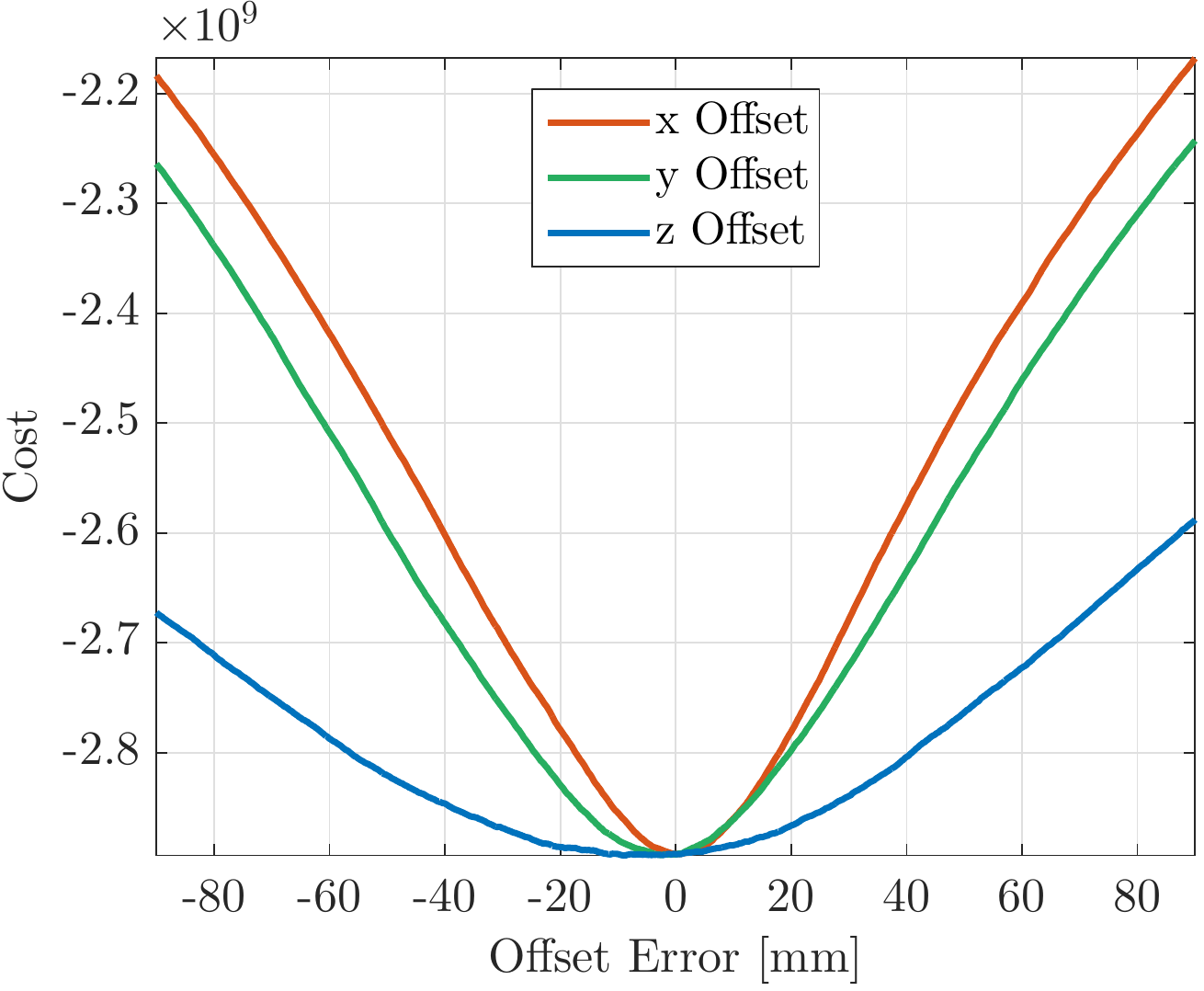}
    \caption{Offset error versus cost for each translational parameter, with others held constant.}
\end{subfigure}
~
\begin{subfigure}[t]{0.3\textwidth}
    \centering
    \includegraphics[width=\textwidth]{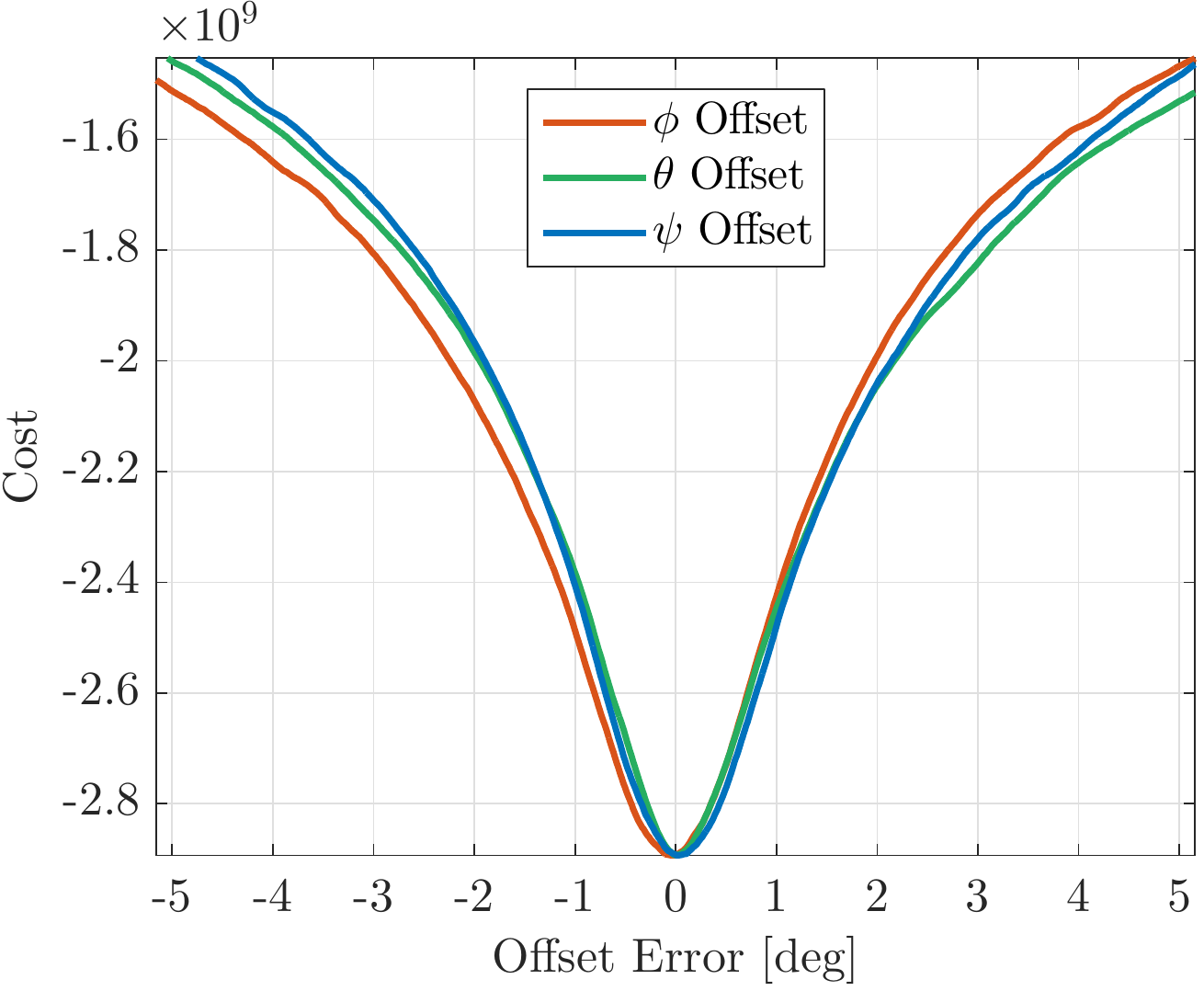}
    \caption{Offset error versus cost for each angular parameter, with others  held constant.}
\end{subfigure}
~
\begin{subfigure}[t]{0.3105\textwidth}    
    \centering
    \includegraphics[width=\textwidth]{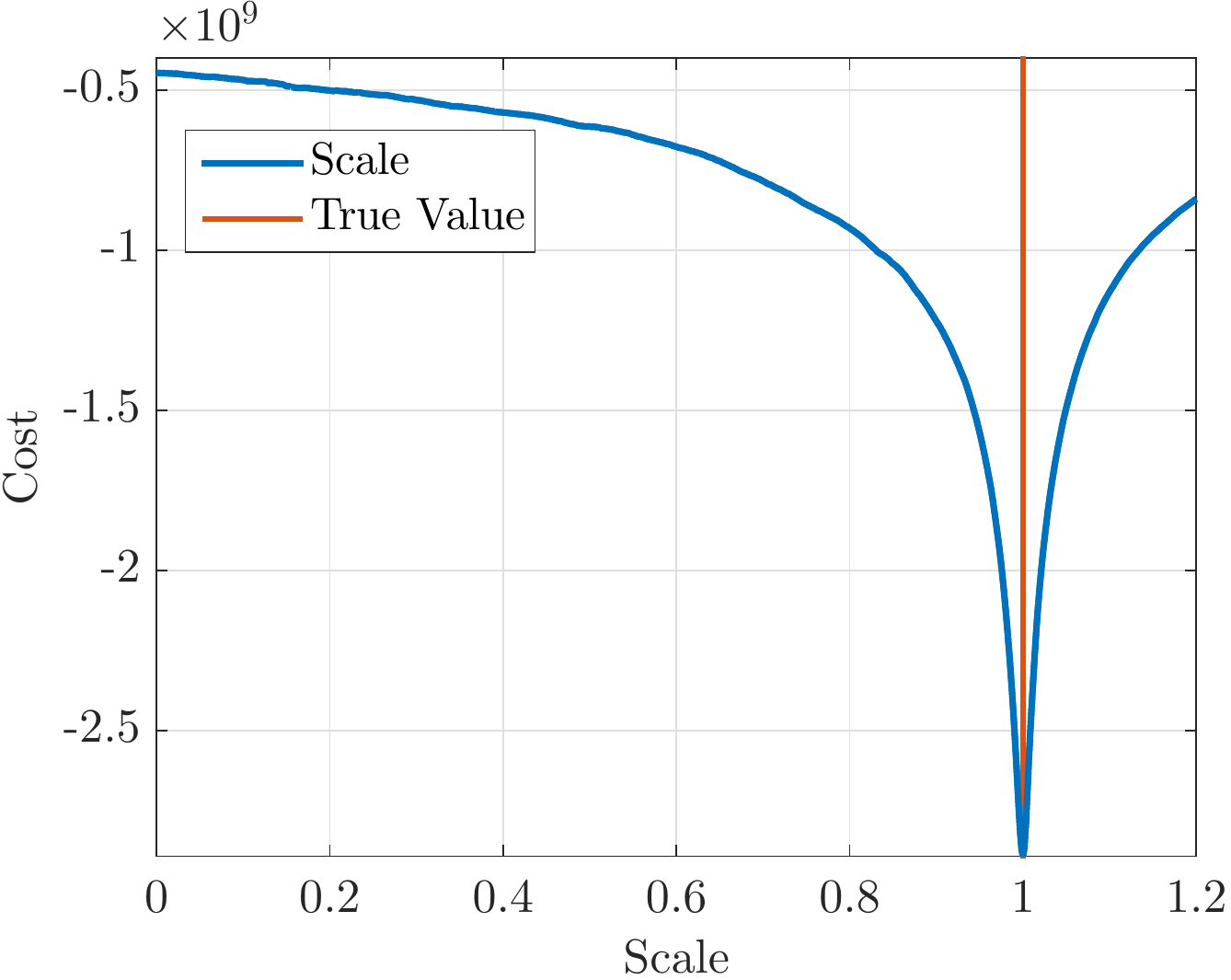}
    \caption{Scale parameter versus cost (blue) with true value at scale $=1$ (red), with $SE{3}$ transform parameters held constant.}
    \label{fig:scale}
\end{subfigure}
\caption{Effect of variation of individual parameters on cost function, with other parameters held fixed at their true values.}
\label{fig:single_params}
\end{figure*}

Entropy minimization is an intuitive way to quantify point cloud crispness and therefore extrinsic calibration accuracy, but the measured entropy is dependent on the sensor trajectory and the environment, so it is difficult to derive convergence guarantees for this approach. Instead, we provide experimental validation. While it is difficult to visualize this high dimensional optimization problem, we can gain some insight by varying each parameter individually while holding the other parameters constant. In \Cref{fig:single_params}, we see that even with noise, the cost function is minimized very close to the calibration parameters' true values. This test was performed with the aforementioned noise values in the \emph{Simple Room} environment. Nevertheless, outside of the region near the true calibration values we have observed that the cost function may not be particularly smooth; we expect the existence of several local minima, increasing the difficulty of the global optimization task.

\subsection{Global Optimization}
\Cref{tab:sim_results} summarizes calibration accuracy for each parameter in each simulation environment by presenting the average absolute error over 10 randomly generated trajectories. Our method achieves millimetre translational and sub-degree rotational accuracy.

In practice, we found that the base sensor trajectory determined algorithm convergence far more than the environment itself. Some individual runs for the more challenging environments, in particular \emph{Quadratic Forest} and \emph{Triangle Array}, produced better results than a typical run in the \emph{Simple Room} environment. In general, the algorithm converges reliably if the lidar repeatedly scans surfaces from several different viewpoints.

Occasionally, the randomly generated trajectories were identified as degenerate and discarded. We found, in particular, that the roll $\phi$ and yaw $\psi$ parameters  were unobservable for several trajectories. We recognize that this is a consequence of our trajectory generation approach, which we intend to improve upon in future work. We posit that the estimation of pitch $\theta$ is a better representation of achievable  accuracy of the angular parameter calibration.

\subsection{Temporal Calibration}\label{sec:sim_td}
Using simulation data, we examined the possibility of adapting our approach for temporal calibration of the sensor data streams, which is often necessary to achieve higher-accuracy results. To this end, we added a 20 ms delay to the lidar scans with respect to the associated camera poses. This represents a worst case scenario based on a state-of-the-art spatiotemporal calibration algorithm \cite{2014_Rehder_Spatio-Temporal}. 
\begin{figure}
\vspace*{-2mm}
    \centering
    \begin{subfigure}[t]{0.8\columnwidth}
\centering
\includegraphics[width=\textwidth]{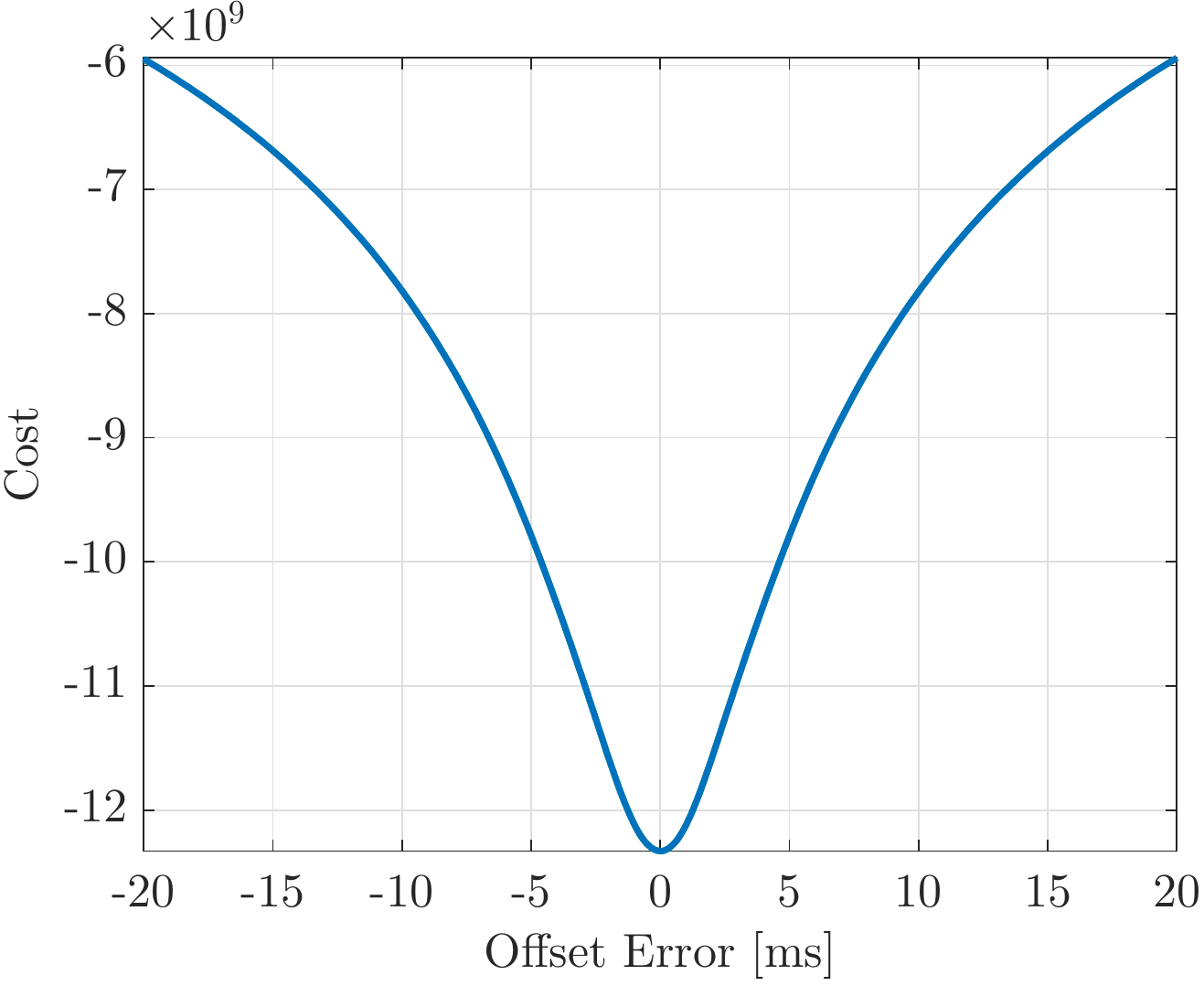}
\caption{Error in temporal offset between sensors when the \textit{Sim}(3) transform parameters are held constant at their true values as shown in \Cref{tab:td_results}.}\label{fig:td_results_true}
    \end{subfigure}
    \begin{subfigure}[t]{0.8\columnwidth}
\centering
\includegraphics[width=\textwidth]{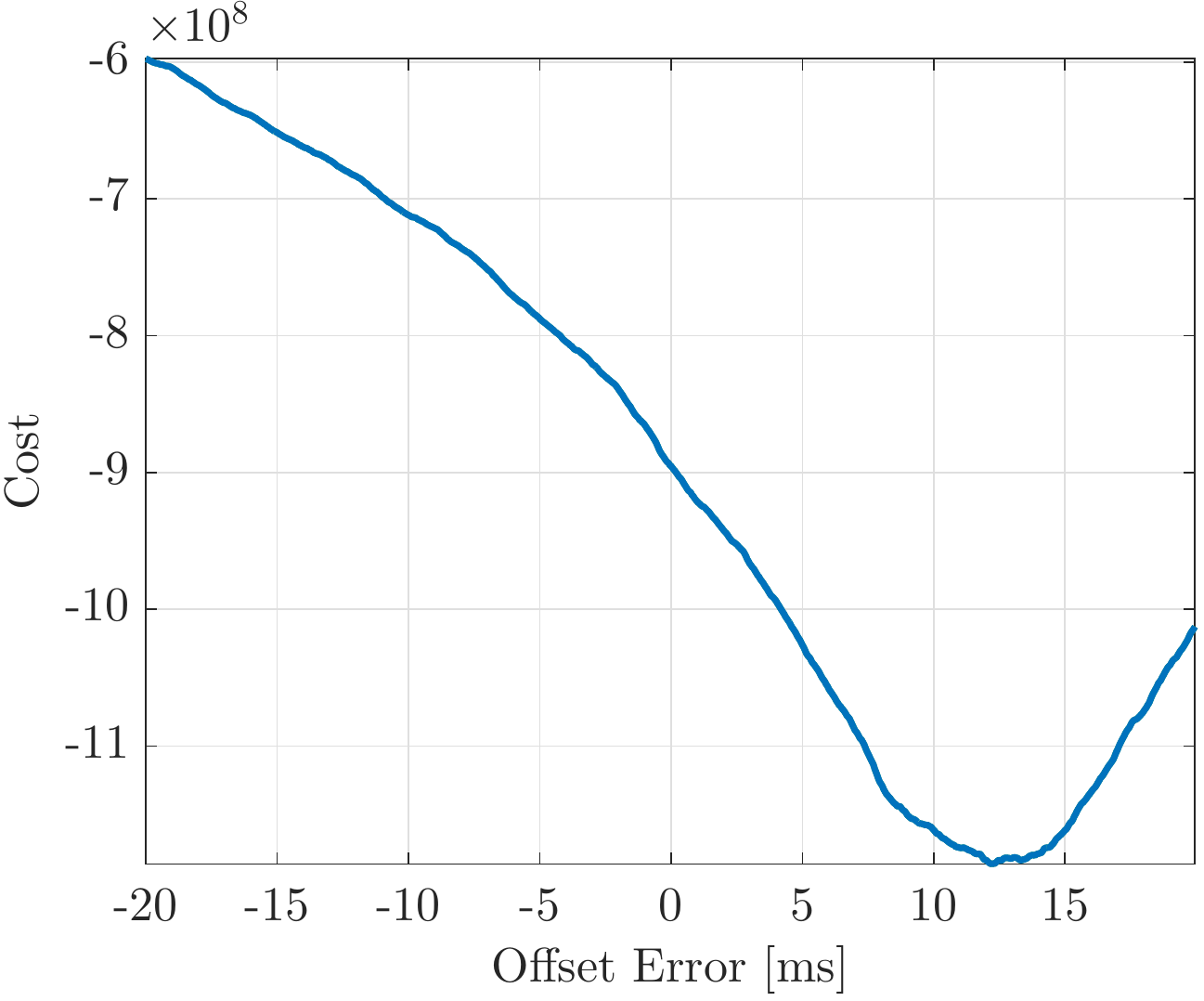}
\caption{Error in temporal offset between sensors when the \textit{Sim}(3) transform parameters are held constant at the seed values shown in \Cref{tab:td_results}.}\label{fig:td_results_false}
    \end{subfigure}
    \caption{Sample result of the temporal pre-calibration approach, based on a trajectory generated in the \emph{Simple Room} environment.}\label{fig:td_results}
\end{figure}

Rather than performing spatial and temporal calibration jointly, we followed \cite{MaddernICRA2012} and instead attempted to perform temporal calibration prior to spatial calibration, again through entropy minimization (instead of using, e.g., the TICSync library \cite{HarrisonTICSync}). We locked the \textit{Sim}(3) transform parameters at nominal values and varied the time delay $t_d$ between the camera and lidar data streams, performing a simple linear interpolation between camera poses. Once the optimal value of $t_d$ was found, we then held that value constant as we carried out a global optimization over the \textit{Sim}(3) parameters. A sample result using the true spatial calibration parameters is shown in \Cref{fig:td_results_true}, demonstrating that for a noisy trajectory in the \emph{Simple Room} environment, it is possible to recover the temporal offset exactly.

In realistic scenarios, however, the spatial transform will not be known exactly initially. In \Cref{tab:td_results}, we show the average time offset error after temporal calibration, when the spatial transform parameters are perturbed from their true values. The parameters were seeded $\pm 30$ mm and approximately $\pm 5^{\circ}$ away from ground truth; this represents an initial guess accuracy that should be easily attainable in practice. \Cref{fig:td_results_false} shows how, given a poor initial guess, the estimated time delay between sensors can be inaccurate.

The results presented in \Cref{tab:td_results} indicate how much more difficult point cloud reconstruction becomes when a temporal offset is unaccounted for. Pre-calibrating the time offset is shown to substantially improve the accuracy of the \textit{Sim}(3) parameter estimates, but the errors remain significantly larger than those in \Cref{tab:sim_results}, especially for the angular values. Given the result illustrated by \Cref{fig:td_results_true}, we hypothesize that simultaneously estimating the temporal and spatial offset between sensors through entropy minimization could prove to be more reliable than a pre-calibration approach.

\begin{table}
\centering
\caption{Average absolute error on 10 temporally delayed data sets in the \emph{Simple Room} environment, with and without temporal pre-calibration.}
\begin{threeparttable}
\begin{tabular}{ c  c  cc  c  c }
 & & && \multicolumn{2}{c}{Average absolute error -- $\mu (\sigma)$} \T\B \\ \cline{5-6}
Parameter & Initial & True && N.T.C.$^1$ & T.C.$^2$ \T\B \\ \hline
$t_d$ [ms]& - & 20.0  && - & 5.6 \T \\
$x$ [mm] & -230 & -200 && 10.7 (7.8) & 3.81 (1.89)\\
$y$ [mm] & 80.0 & 50.0 && 7.1 (6.5) & 3.0 (2.2)\\
$z$ [mm] & 330 & 300 && 18.2 (10.2) & 4.9 (3.5)\\
$\phi$ [deg] & 9.74 & 14.3 && 0.67 (0.59) & 0.39 (0.42)\\
$\theta$ [deg] & 91.7 & 97.4 && 0.15 (0.13) & 0.10 (0.06)\\
$\psi$ [deg] & 63.0 & 57.3 && 0.79 (0.55) & 0.47 (0.40) \\
 Scale & 1.2 & 1.0 && 2.0 (1.2)$\times 10^{-3}$& 0.81 (0.52)$\times 10^{-3}$\B \\
\hline
\end{tabular}
\begin{tablenotes}
\item[1] Not Temporally Calibrated
\item[2] Temporally Calibrated, with average temporal delay ($t_d$) value over 10 runs of 14.4 ms (true value was 20 ms).
\end{tablenotes}
\end{threeparttable}\label{tab:td_results}
\end{table}

\section{EXPERIMENTS}\label{sec:exp}

\subsection{Experimental Setup}\label{sec:setup}
\begin{figure}
    \centering
    \begin{subfigure}[t]{0.45\textwidth}
    \centering
        \includegraphics[width=0.6\columnwidth]{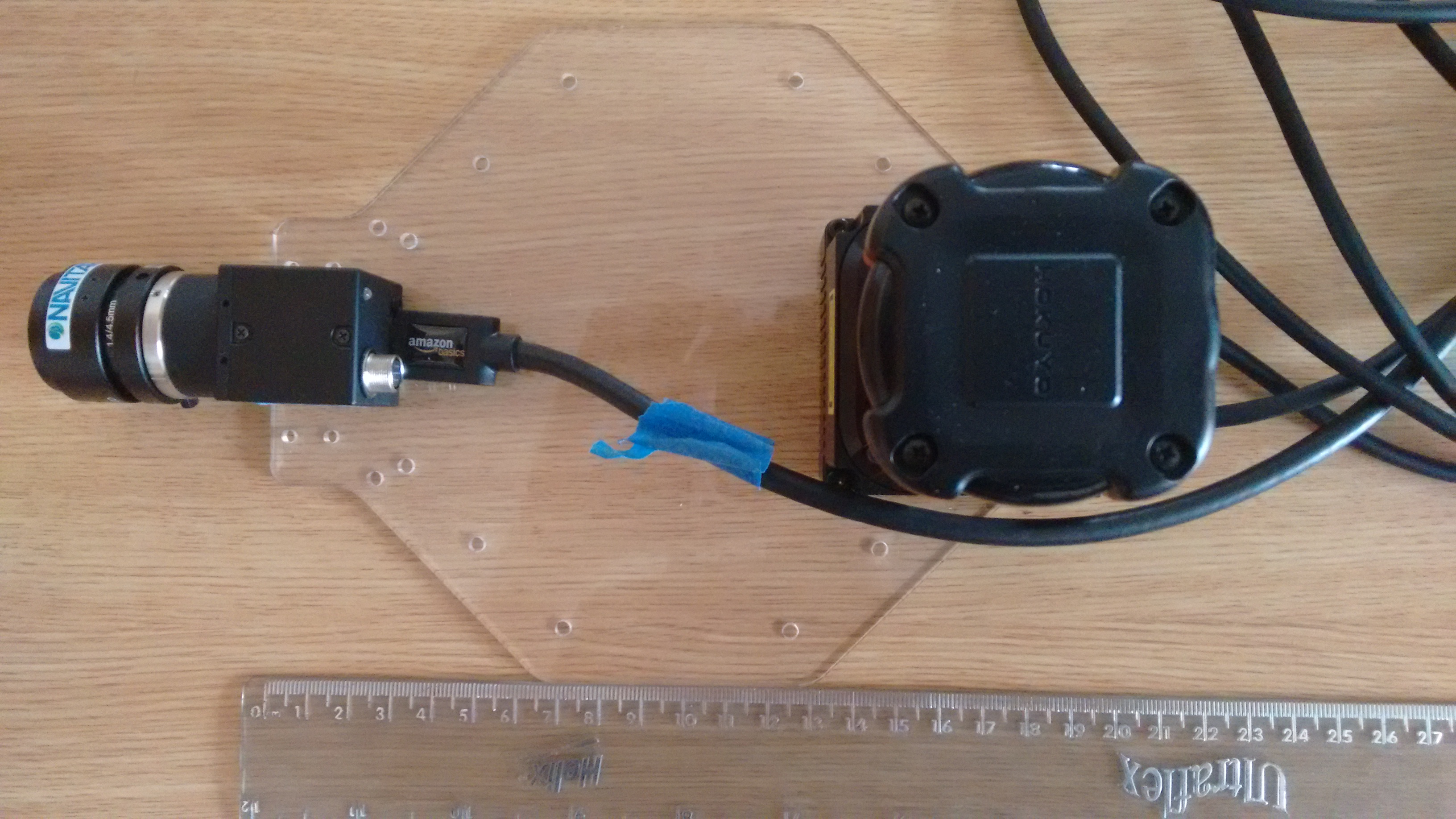}
        \caption{Configuration 1: with overlapping sensor FOV.} \vspace{7pt}
    \end{subfigure}
    \centering
    \begin{subfigure}[t]{0.45\textwidth}
     \centering
     \includegraphics[width=0.6\columnwidth]{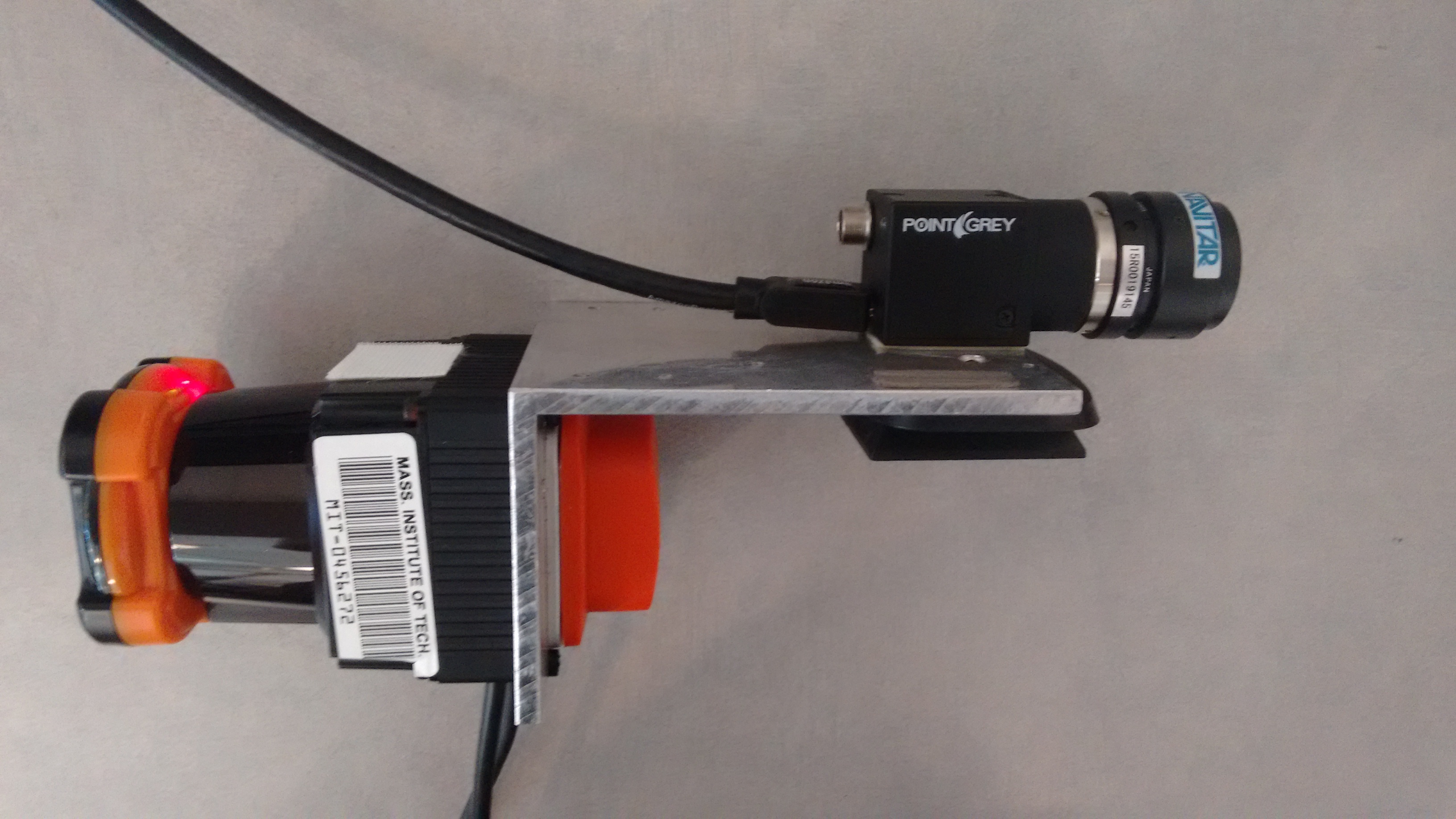}
     \caption{Configuration 2: non-overlapping sensor FOV.}
    \end{subfigure}
    \caption{Hardware configurations used in experiments.}\label{fig:sensor_rigs}
    \vspace{-2mm}
\end{figure}

Hardware experiments were conducted with a hand-held sensor rig consisting of a Hokuyo UTM-30LX laser rangefinder and a PointGrey Flea3 monocular camera. The sensors were rigidly fixed to two separate mounts displayed in \Cref{fig:sensor_rigs}a and \Cref{fig:sensor_rigs}b. Data were collected via USB  on a laptop configured with ROS \cite{Quigley2009}. All laser data was collected at 40~Hz and 0.25~degree angular resolution, while the camera data was collected at 200~fps with a resolution of 640 $\times$ 512 pixels per frame. We estimated the temporal offset of the lidar relative to the ROS clock using a built-in procedure, then paired the camera and lidar messages according to their timestamps to obtain an approximately synchronized 30~Hz data stream. Given the synchronized data, we estimated the trajectory of the camera (up to an unknown scale factor) using ORB-SLAM2 \cite{MurArtal:2015:TRO}.\footnote{\url{https://github.com/raulmur/ORB_SLAM2}}

\begin{figure}
\centering
\includegraphics[width=0.8\columnwidth]{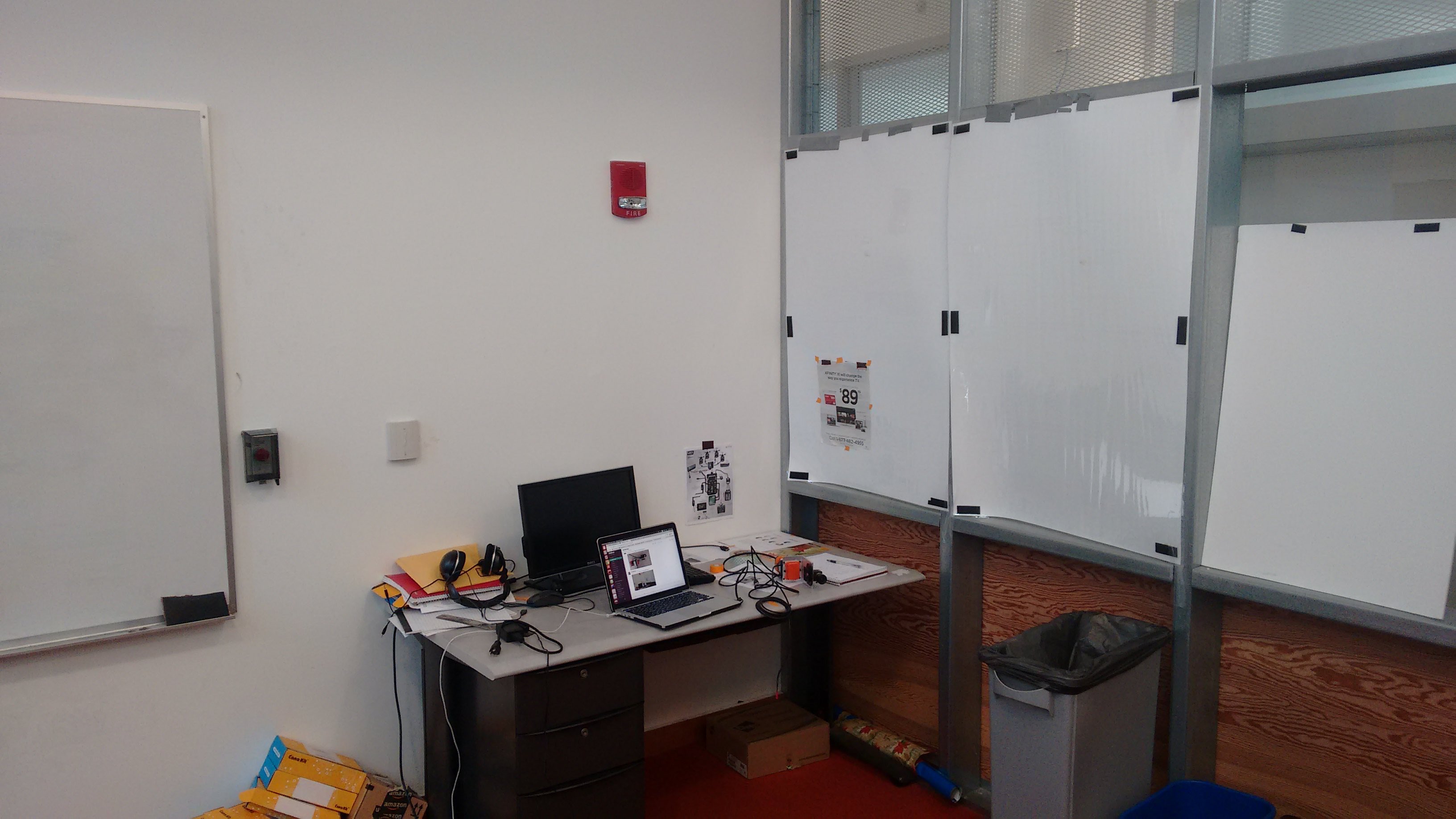}
\caption{Cluttered office space in the MIT's Stata Center used for data collection.}\label{fig:lids_environment}
\end{figure}

Several data sets were collected in the office space at MIT's Stata Center, shown in \Cref{fig:lids_environment}. The sensor rig was excited manually, taking care not to capture images or laser measurements of the operator's body.

\subsection{Global Optimization}
Calibration results are shown in \Cref{tab:real_results}. A data set was collected for each of the sensor configurations shown in \Cref{fig:sensor_rigs}. Individual data sets contain 2-4 minutes of lidar and camera measurements. To keep the optimization tractable, we split the data sets into segments containing at most 1.9 million lidar points, as in the simulations. 
For each data set, we initialized the scale parameter by roughly estimating a bounding box around the trajectory during data collection, then comparing with the ORB-SLAM2 trajectory. The scale factor is set based on the choice of keyframes used during ORB-SLAM2 initialization, and hence varies per data set.
\begin{table*}
    \centering
    \caption{Calibration results for the \textbf{overlapping} case, for one data set segmented into four trial runs.}
        \begin{tabular}{*{9}{c}}
            && \multicolumn{7}{c}{Calibration results} \T\B \\ \cline{3-9}
             && $x$ $\left[\mathrm{mm}\right]$ & $y$ $\left[\mathrm{mm}\right]$ & $z$ $\left[\mathrm{mm}\right]$ & $\phi$ [deg] & $\theta$ [deg] & $\psi$ [deg] & Scale $[\times 10^{-3}]$ \T\B \\ \hline
            \textbf{Initial Guess} && $\mbf{160.0}$ & $\mbf{0.0}$ & \textbf{-}$\mbf{50.0}$ & $\mbf{0.400}$ & \textbf{-}$\mbf{90.00}$& $\mbf{0.00}$ & \textbf{-}$\mbf{90.00}$  \T\B \\ 
	    Trial I && -178.2 & -3.8 & -45.8  & 90.58 & -0.10 & -90.74 & 0.506  \\
	    Trial II && 182.5 & -2.8 & -50.9  & -90.22 &  0.14 & -90.12 & 0.509 \\
	    Trial III && 173.6 & -2.9 & -47.8  & 89.84 & -0.27 & -90.04 & 0.506 \\
	    Trial IV && 187.0 & -4.9 & -54.2  & -89.49 & -0.02 & -90.27 & 0.511 \B \\ \hline
	    $\mu$ $\left( \sigma \right)$ && 180.3 (5.0) & -3.6 (0.8) & -49.7 (3.2)  & -90.03 (0.41) & 0.06 (0.15) & -90.29 (0.27) & 0.508 (0.002) \T \B \\ \hline 
        \end{tabular}
    \label{tab:real_results}
\end{table*}
\begin{table*}
    \centering
    \caption{Calibration results for the \textbf{non-overlapping} case, for one data set segmented into three trial runs.}
        \begin{tabular}{*{9}{c}}
            && \multicolumn{7}{c}{Calibration results} \T\B \\ \cline{3-9}
            && $x$ $\left[\mathrm{mm}\right]$ & $y$ $\left[\mathrm{mm}\right]$ & $z$ $\left[\mathrm{mm}\right]$ & $\phi$ [deg] & $\theta$ [deg] & $\psi$ [deg] & Scale $[\times 10^{-3}]$ \T\B \\ \hline
                        \textbf{Initial Guess} && $\mbf{50.0}$ & $\mbf{0.0}$ & \textbf{-}$\mbf{250.0}$ & $\mbf{180.00}$ & $\mbf{0.00}$ & \textbf{-}$\mbf{90.00}$ & $\mbf{0.500}$ \T\B \\
	      Trial I && 45.2 & 0.5 & -202.1 & 180.13 & -1.11 & -88.89 & 0.2149 \\
	      Trial II && 42.2 &-1.2 & -202.8 & 180.73 & -1.29 & -88.93 & 0.2159 \\
	      Trial III && 44.0 & 0.2 & -204.2 & 180.43 & -1.78 & -88.89 & 0.2156 \B \\ \hline
	    $\mu$ $\left( \sigma \right)$   && 43.8 (1.2) & -0.2 (0.8) & 203.0 (0.85) & 180.43 (0.24) & -1.39 (0.28) & -88.90 (0.02) & 0.2156 (0.0005)\T \B \\ \hline
        \end{tabular}
    \label{tab:real_results_non}
    \vspace{-2mm}
\end{table*}

The \textit{Sim}(3) transform parameters recovered in our experiments are consistent between runs, with very few outliers. We note, however, that the data sets in \Cref{tab:real_results} and \ref{tab:real_results_non} feature smooth, slow motion in order to minimize the effect of inaccurate temporal calibration, which we found to be critical to obtaining reliable results. Furthermore, the current optimization procedure is computationally expensive for larger data sets, forcing us to significantly reduce the $k$ parameter in Equation \ref{eq:kfactor}, and consequently decreasing the accuracy of the cost function\cite{MaddernICRA2012}. We are actively exploring alternative optimization procedures that would make the computation more tractable, which we believe would lead to more precise results. 

\section{CONCLUSION AND FUTURE WORK}\label{sec:conc}
The automatic approach presented in this paper represents a generalized method for 2D lidar extrinsic calibration. While we focus on 2D lidar to monocular camera calibration, the technique could just as effectively be used to calibrate a 2D, or even 3D lidar to another motion estimation system. GNSS platforms, visual-inertial systems, stereo cameras, and 3D lidars are just a few examples of systems that could make use of the proposed approach---we intend to explore these combinations in future work. The distinct value of this calibration procedure is that it can be performed in virtually any environment, with no limitations on sensor configuration.

We have found in practice that the mis-estimation of the temporal offset between the sensor data streams can severely degrade the quality of the \textit{Sim}(3) transform parameter estimates. Our temporal pre-calibration approach has limited accuracy, but the results presented in this work motivate the use of RQE minimization in general spatiotemporal calibration. 

Finally, we intend to improve the computational tractability of our algorithm to enable online calibration. In particular, we intend to exploit modern GPGPU processing to parallelize the pairwise entropy computations in the cost function, allowing for faster and potentially more accurate calibration.

\section*{ACKNOWLEDGMENTS}

The authors would like to thank Kyel Ok from MIT CSAIL's Robust Robotics Group for his invaluable help in collecting the data sets used in our experiments.

\bibliographystyle{IEEEtran}
\bibliography{mfi2016}

\end{document}